%% file: main.tex
\documentclass{article}

\usepackage{microtype}
\usepackage{graphicx}
\usepackage{subfigure}
\usepackage{booktabs} % for professional tables
\usepackage{amsmath,amssymb,amsfonts}
\usepackage{algorithmic}
\usepackage{textcomp}
\usepackage{xcolor}
\usepackage{dblfloatfix}

\usepackage{hyperref}

\usepackage[accepted]{icml2021}

\input{header}

\newcommand{\Tau}{\mathcal{T}}

\newcommand{\method}{GateL0RD\xspace}

\icmltitlerunning{Developing hierarchical anticipations via neural network-based event segmentation}

\begin{document}

\twocolumn[
\icmltitle{Developing hierarchical anticipations \\ via neural network-based event segmentation}

\begin{icmlauthorlist}
\icmlauthor{Christian Gumbsch}{tue,mpi}
\icmlauthor{Maurits Adam}{pot}
\icmlauthor{Birgit Elsner}{pot}
\icmlauthor{Georg Martius}{mpi}
\icmlauthor{Martin V. Butz}{tue}
\end{icmlauthorlist}

\icmlaffiliation{tue}{Neuro-Cognitive Modeling Group, University of Tübingen, Tübingen, Germany}
\icmlaffiliation{mpi}{Autonomous Learning Group, Max-Planck Institute for Intelligent Systems, Tübingen, Germany}
\icmlaffiliation{pot}{Developmental Psychology, University of Potsdam, Potsdam, Germany}

\icmlcorrespondingauthor{Christian Gumbsch}{christian.gumbsch@uni-tuebingen.de}

\icmlkeywords{developmental robotics, goal-predictive gaze, event cognition, temporal abstractions}

\vskip 0.3in
]

\printAffiliationsAndNotice{}%\icmlEqualContribution} % otherwise use the standard text.

\begin{abstract}
Humans can make predictions on various time scales and hierarchical levels.
Thereby, the learning of event encodings seems to play a crucial role. 
In this work we model the development of hierarchical predictions via autonomously learned latent event codes.
We present a hierarchical recurrent neural network architecture, whose inductive learning biases foster the development of sparsely changing latent state that compress sensorimotor sequences. 
A higher level network learns to predict the situations in which the latent states tend to change.
Using a simulated robotic manipulator, we demonstrate that the system (i) learns latent states that accurately reflect the event structure of the data, (ii) develops meaningful temporal abstract predictions on the higher level, and (iii) generates goal-anticipatory behavior similar to gaze behavior found in eye-tracking studies with infants.
The architecture offers a step towards the autonomous learning of compressed hierarchical encodings of gathered experiences and the exploitation of these encodings to generate adaptive behavior. 
\end{abstract}

\section{Introduction}
Humans are able to generate hierarchical predictions on various time scales.
How does this ability develop?
Developmental psychology has shown that goal-predictive eye gaze develops during the first year of life: Infants start to look at the goal of an action before it is concluded.
The generation of this goal-anticipatory gaze depends on various factors such as agent familiarity, 
action familiarity, the competence to execute the action themselves, agency cues, and action effects \cite{Adam2016,Adam2018,Adam2020,AdamClaw2021,elsner2021infants,Gredeback2010infants,Kanakogi2011}. 

\begin{figure}
    \startsubfig{}
    
  \begin{tabular}{@{}c@{\ \ }c@{\ \ }}
   \includegraphics[width=0.5\linewidth]{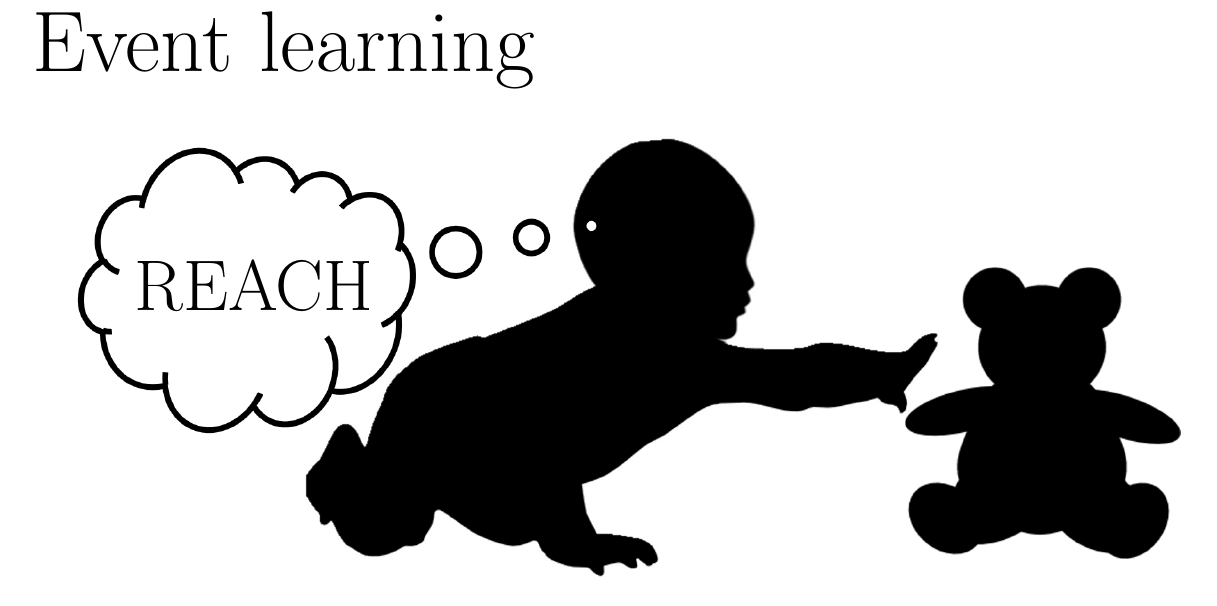}&
     \includegraphics[width=0.5\linewidth]{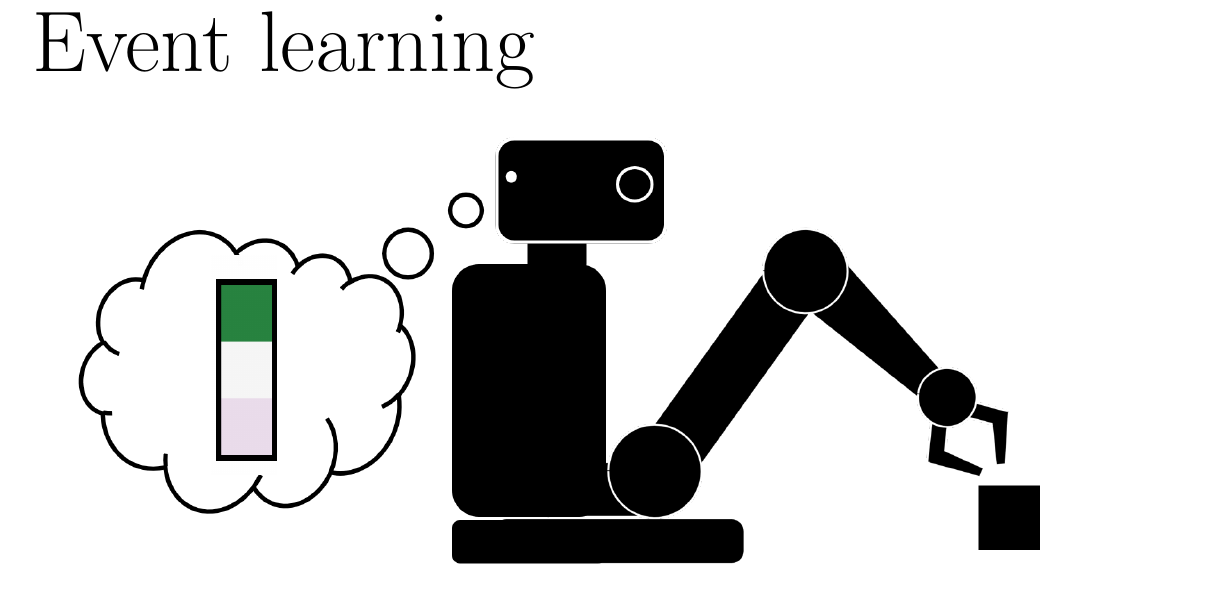}\\[-0.3em]
     \includegraphics[width=0.5\linewidth]{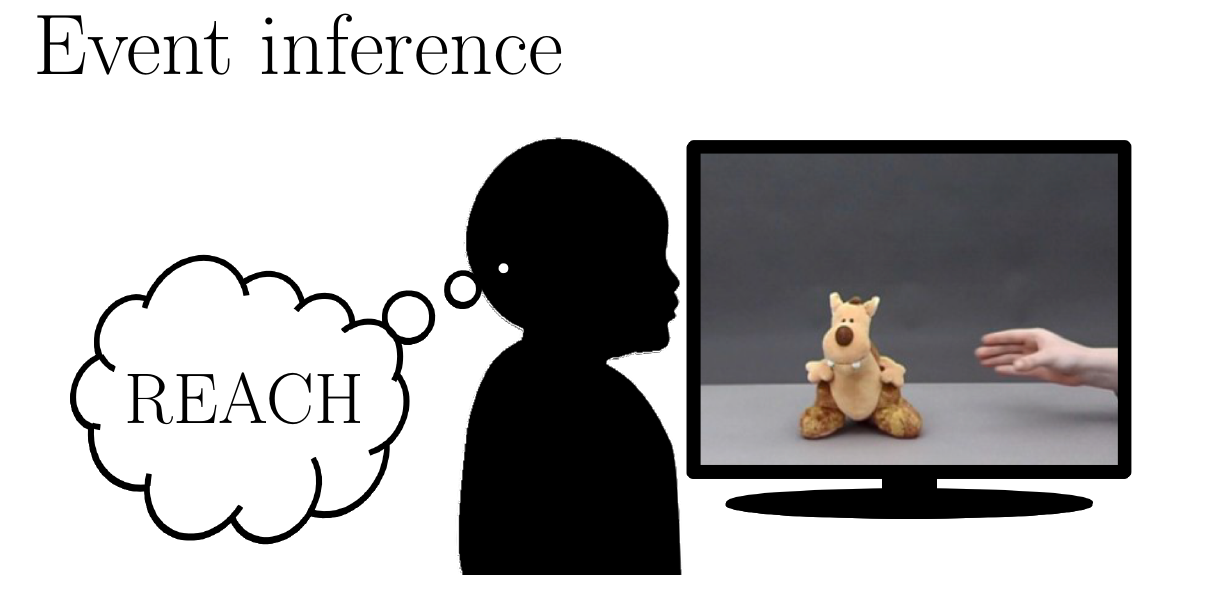}&
     \includegraphics[width=0.5\linewidth]{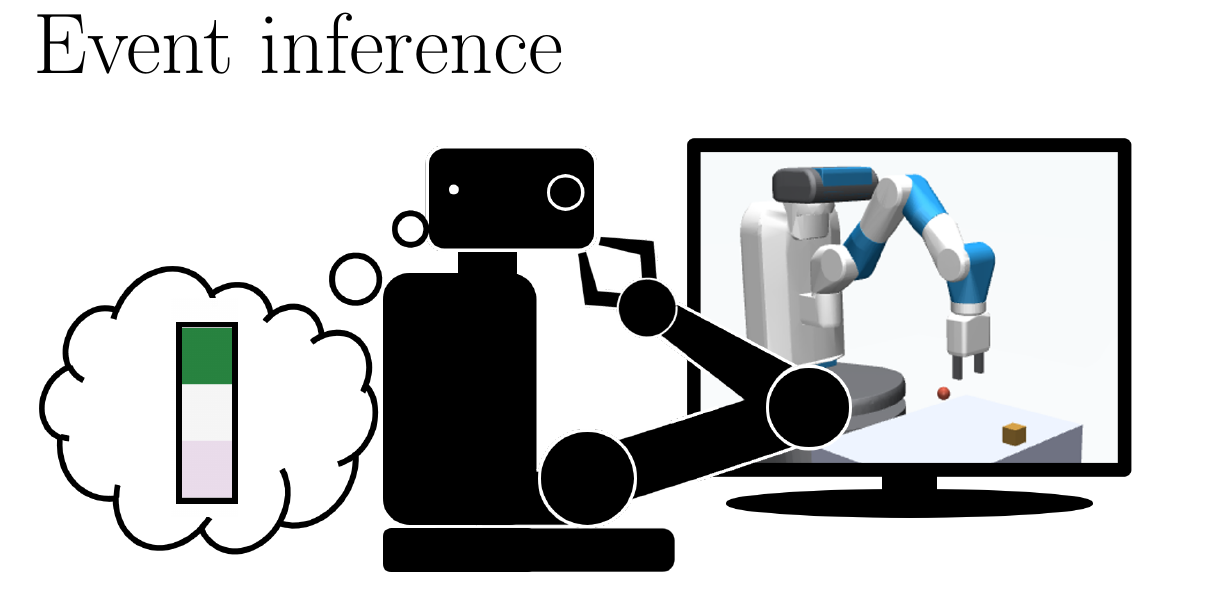}\\[-0.3em]
     \includegraphics[width=0.5\linewidth]{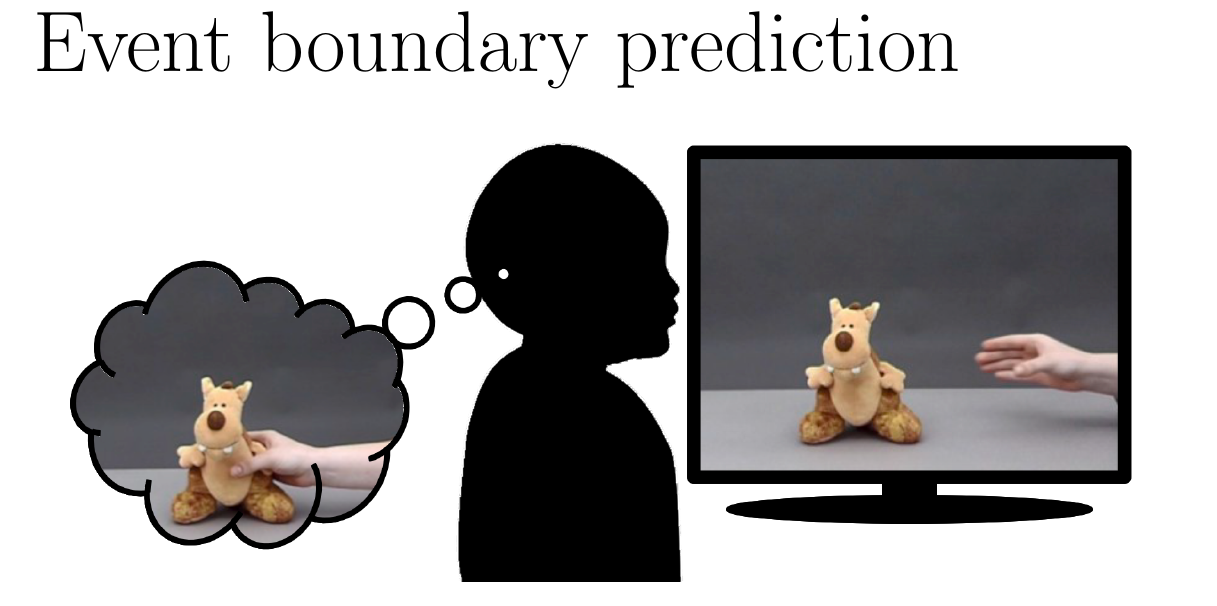}&
     \includegraphics[width=0.5\linewidth]{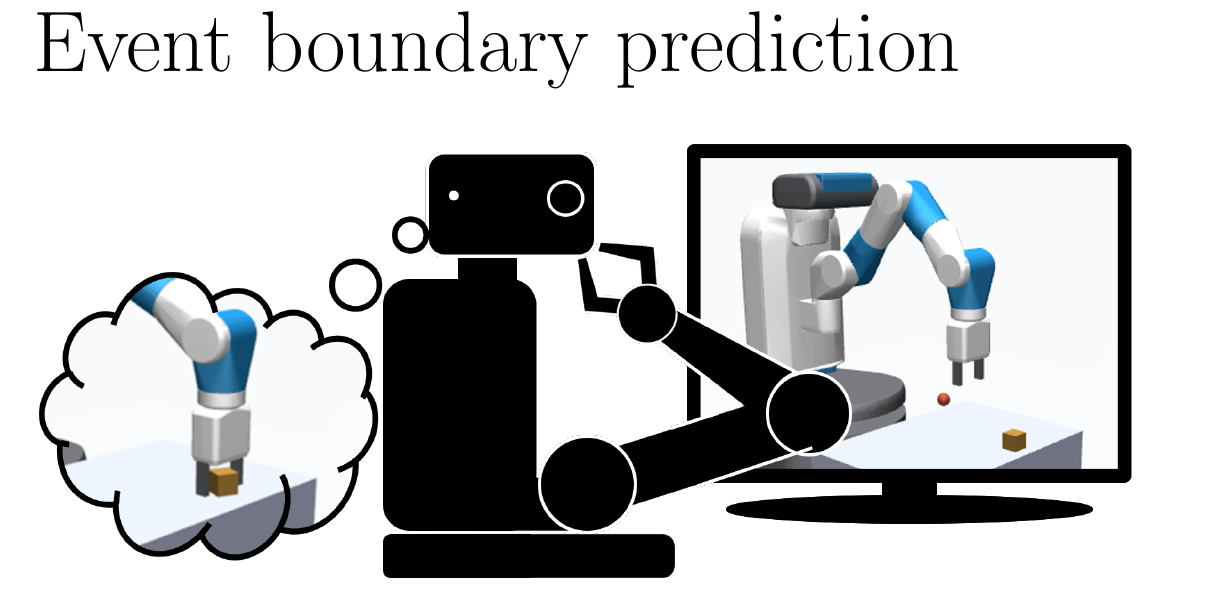}\\
     \subfig{fig:hypo1} & \subfig{fig:hypo2}\\[-0.3em]
  \end{tabular}\vspace*{-0.02cm}
  \caption{Hypothesis for modeling infants' goal anticipations. (a): Infants first learn event representations (top). When they see familiar movements, these representations get re-activated (middle), enabling the anticipation of action consequences (bottom). (b): We implement this process in artificial systems, which learn event codes in a self-supervised manner. 
  Figure is adapted from \cite{CAPRI} with screenshots from \cite{Adam2020}.} \label{fig:hypothesis}
\end{figure}

Theories on event-predictive cognition \cite{Butz:2016,Butz2021,EST} have led to the development of a theoretical framework, which attempts to explain these findings 
\cite{elsner2021infants} and which is illustrated in \fig{fig:hypothesis}. 
The main assumption is that infants learn to compress their action experiences into action-event schemata. 
These schemata encode perceivable features of the entities, their movements and the goal.
Once learned, the observation of a known event, \eg, a hand moving towards an object, results in the activation of the matching event schema.
This activated schema leads to the generation of goal-predictive gaze shifts, striving to minimize uncertainty about the predicted end of the event, \eg, the hand touching the object.

Elsewhere \cite{CAPRI}, the development of such goal-predictive gaze shifts was modeled via performing active inference, \ie predicted uncertainty minimizing behavior, on the level of events. 
First, event schemata are learned. 
Then, aiming at minimizing anticipated uncertainty within an event and between events, goal-predictive eye gaze develops mimicking infant behavior. 
However, the existing implementation has various restrictions: 
It learns event models based on supervised segmentations and requires fully-observable, linear simulations.
This prevents the application of the model to more realistic scenarios, where motions are boundedly complex but non-linear, information can be hidden, and events are unknown. 

Here we introduce a hierarchical recurrent neural network architecture, which can be trained end-to-end.
The architecture learns to compress sensorimotor sequences into sparsely changing latent states. 
Based on these compressions, it learns temporal abstract predictions to anticipate the observations encountered upon latent state changes.
Striving to minimize uncertainty, the system develops goal-predictive behavior similar to the way it develops in infants.

\section{Event cognition}

We humans are exposed to a \emph{continuous} stream of perceptual information.
However, a large body of psychological, neurological, and linguistic evidence suggests that we perceive, memorize, and predict information in terms of \emph{discrete} events \cite{EST, ZacksTversky, ZacksBook, Butz2021}.
According to Event Segmentation Theory (EST) \cite{EST}, and in line with various  theories on predictive processing \cite{FEP, Hohwy, Clark}, humans continuously attempt to predict future perceptual input.
EST argues that event encodings develop to better predict the next immediate perceptions as well as to anticipate event successions \cite{Butz:2016,EST, ZacksBook}. 
In fact, the resulting hierarchical structures may constitute the foundations for abstract thought, counterfactual reasoning, and versatile planning \cite{Butz2021, Eppe2022, Botvinick2009}.
Meanwhile, the Theory of Event Coding (TEC) \cite{TEC} argues that humans encode both actions and perceptions in event codes, emphasizing the close connection between action and perception.

Using these insights to develop event representations within an artificial system is not straight forward. 
Recurrent neural networks (RNNs) are commonly used in deep learning to process sensorimotor time series.
However, they encode experience in a fundamentally different way:
Namely, they update their latent representation of the ongoing activity in every time step.
More abstract latent codes can emerge in hierarchical RNNs with levels that operate on different time scales \cite{tani2016exploring}. 
Previously, this was implemented by continuously updating higher level layers at a slower, predefined update rate, \eg \cite{MTRNN, ClockworkRNN}.
In contrast to that, theories on event cognition \cite{EST, ZacksTversky, ZacksBook, Butz2021} suggest that event encodings are mostly updated upon encountering critical situations, \ie \emph{event boundaries}, not based on fixed time scales. 
Recently, Gumbsch et al.\ \cite{GateL0RD}, proposed Gated L0-Regularized Dynamics (\method), an RNN that only sparsely updates its latent state when the task demands it.
In this work, we demonstrate that embedding \method in a suitable sensorimotor learning architecture fosters the self-supervised development of latent event codes.

\section{Hierarchical event-inference architecture}
Our architecture learns to process time series with sparsely changing latent states, which modulate predictive forward-inverse models.
It furthermore learns to predict situations when the latent states tend to change.
These predictive abilities enable goal-anticipatory behavior.
We now detail our architecture and show how its inductive biases enable it to develop (i) event-compressing latent states, (ii) the prediction of situations for which the latent codes are expected to change, and (iii) goal-anticipatory behavior.

\subsection{Sparsely changing memory}

\begin{figure}
    \startsubfig{}
  \begin{tabular}{@{}c@{\ \ }c@{\ \ }}
  \subfig{fig:gatel0rd_components} & \subfig{fig:retanh}\\[-1.3em]
   \includegraphics[width=0.64\linewidth]{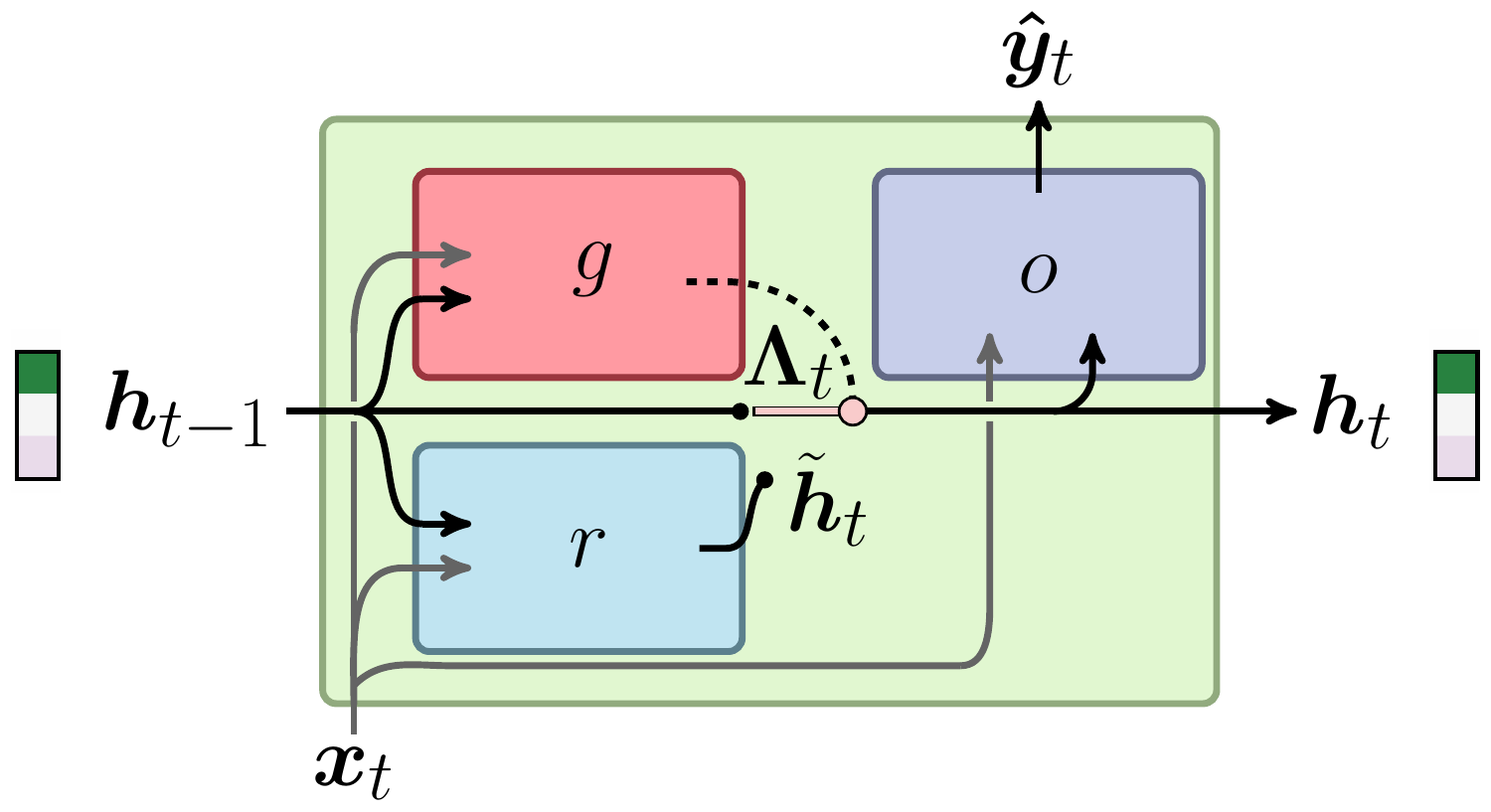} & \raisebox{0.5cm}{\includegraphics[width=0.33\linewidth]{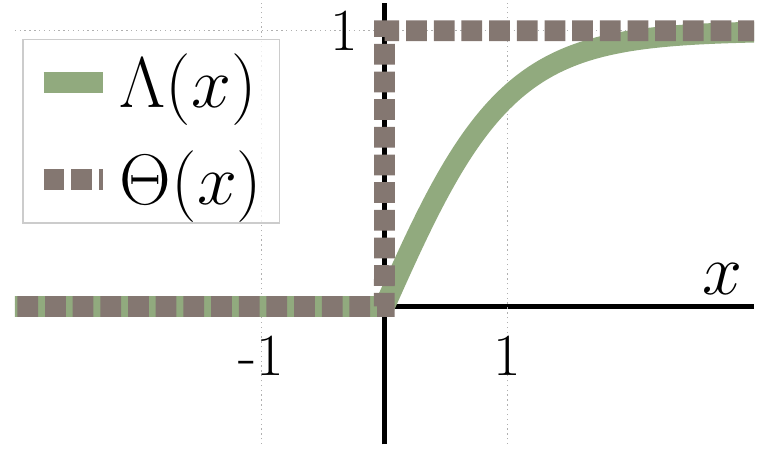}}\\[-0.3em]
     
  \end{tabular}\vspace*{-0.02cm}
    \caption{\method \cite{GateL0RD}: (a) Internal components, (b) its update gate activation  $\Lambda$ and binarization via the step function  $\Theta$.}
    \label{fig:gatel0rd}
\end{figure}

We describe the environment by a partially observable Markov decision process (POMDP) with state, action, and observation spaces $\mathcal{S,A,O}$, and hidden transitions $\mathcal S\times \mathcal A \to \mathcal S$.
Our system encodes the unobservable information at time $t$ by means of a latent state $\bm h_t \in \mathbb{R}^H$.\footnote{\textbf{Notation:} We denote vectors with bold lowercase letters (\eg $\bm x$) and matrices as bold uppercase letters (\eg $\bm W$). 
Vector dimensions are denoted by superscript and additional information by subscript (\eg $\bm x_t = [x_t^1, ..., x_t^n]$). 
} To learn the latent states $\bm h_t$ we use \method \cite{GateL0RD}. 
Similar to other RNNs, \method implements a general mapping $(\bm{\hat{y}}_t, \bm h_t) = f_{\bm \theta}(\bm x_t, \bm h_{t-1})$, with inputs $\bm x \in \mathbb{R}^N$, outputs $\bm y \in \mathbb{R}^M$, and learnable parameters $\bm \theta$.
Additionally, though, GateL0RD's modularization and a regularizing loss term yield the inductive bias to develop piecewise constant latent states $\bm{h}_t$.

\method's internal structure is sketched-out in \fig{fig:gatel0rd_components}. 
It uses three functions, or subnetworks, a \emph{recommendation function} $r$, a \emph{gating function} $g$, and an \emph{output function} $o$, which together route sensorimotor prediction dynamics and maintain an inner latent state $\bm{h}_{t}$, as follows:
\begin{align}
\hat{\bm{h}}_t &= r(\bm x_t, \bm{h}_{t-1}) \label{eqn:h-proposal}\\
\bm \Lambda_t &= \max(\bm{0}, \tanh( g(\bm x_t, \bm h_t)))\label{eqn:update-gate}\\
\bm{h}_t &= \bm \Lambda_t \odot \hat{\bm h}_t + (\bm 1 - \bm{\Lambda}_t) \odot \bm h_{t-1},\label{eqn:latent-update}\\
\hat{\bm y}_t &= o(\bm x_t, \bm h_t) \label{eqn:network-output}
\end{align}
The recommendation function $r$ proposes a new latent state $\hat{\bm{h}}_t$ (\eqn{eqn:h-proposal}).
The gating function $g$ computes an \emph{update gate} $\bm \Lambda_t \in [0, 1]$ (\eqn{eqn:update-gate}) with a rectified $\tanh$ activation function ($\Lambda$ shown in \fig{fig:retanh}). 
If $\Lambda^i_t > 0$  the latent state is updated at dimension $i$ (\eqn{eqn:latent-update}).
The network output is then determined using the output function $o$ (\eqn{eqn:network-output}). 
To enforce sparse latent state updates, \method is trained using the loss:
\begin{equation}
    \mathcal{L}_{\bm \theta} = \mathbb{E} \Big[ \sum_t \mathcal{L}_{\mathrm{task}} (\bm{\hat{y}}_{t}, \bm{y}_{t})  + \lambda \Theta (\bm{\Lambda}_t) \Big], \label{eq_gatel0rdloss}
\end{equation}
where $\mathcal L_{\mathrm{task}}$ is the task-based loss, \eg MSE and $\Theta$ is the Heaviside step function (shown in \fig{fig:retanh}).
$\Theta (\bm{\Lambda}_t)$ punishes latent state changes and thus fosters the tendency to develop stable, piecewise constant latent states $\bm{h}_t$.
The hyperparameter $\lambda$ modifies the influence strength of the regularization term.

\subsection{Forward-inverse model}
We use \method as a memory module, which is embedded into a predictive forward-inverse model structure as shown in \fig{fig:skip_net}. 
In every time step the model receives an observation $\bm o_t$ and an action $\bm a_t$ as its input. 
At time $t$ the sensorimotor inputs $(\bm a _t, \bm o_t)$ are fed into \method and a network output $\hat{\bm{y}}_t$ and a new latent state $\bm h_t$ are computed as $(\hat{\bm{y}}_t, \bm h_t) = f_\theta(\bm o_t, \bm a_t, \bm h_{t-1})$.
The network outputs $\hat{\bm{y}}_t$ are processed by a forward model $f_{\mathrm{FM}}$, which predicts the next sensory observation $\hat{\bm o}_{t+1}$: 
\begin{equation}
\hat{\bm o}_{t+1} =  f_{\mathrm{FM}}(\hat{\bm y}_{t}).
\label{eq:fm}
\end{equation}
Based on the updated latent state $\bm h_{t+1}$ and the next observation $\bm o_{t+1}$, an inverse model $f_{\mathrm{IM}}$ predicts the next action $\hat{\bm a}_{t+1}$:
\begin{equation}
    \hat{\bm a}_{t+1} = f_{\mathrm{IM}}(\bm o_{t+1}, \bm h_{t+1}).
\label{eq:im}
\end{equation}
\method's initial latent state $\bm h_0$ is determined by an initialization model $f_{\mathrm{init}}$ based on the first sensorimotor inputs:
\begin{equation}
    \bm h_{0} = f_{\mathrm{init}}(\bm a_{1}, \bm o_{1}).
\end{equation}

Our system implements a fully predictive sensorimotor model, that attempts to constantly predict its next sensory input as well as the next agentive action.
However, in many situations the prediction accuracy varies drastically across states of the environment or the agent.
For example, depending on the agent's gaze, different parts of the observation may be noisy or focused.
Similarly, while in some state the model can be very certain about the agent's next actions, in other states a variety of potential actions could follow.
Thus, we want the sensorimotor predictions of our network to be probabilistic, reflecting the heteroscedastic uncertainty in the world.
Instead of directly predicting the next observation $\hat{\bm o}_{t+1}$, the forward model $f_{\mathrm{FM}}$ predicts a probability distribution $p_{\bm \theta}(\hat{\bm o}_{t+1}| \bm o_t, \bm a_t, \bm h_t)$, from which an observation prediction $\hat{\bm o}_{t+1}$ is sampled.
We model $p_{\bm \theta}(\hat{\bm o}_{t+1}| \bm o_t, \bm a_t, \bm h_t)$ as a normal distribution with
\begin{equation}
    p_{\bm \theta}(\hat{\bm o}_{t+1}| \bm o_t, \bm a_t, \bm h_t) = \mathcal{N}(\bm o_t + \bm{\mu}_{t, \Delta o},  \bm{\Sigma}_{t, o}).
\end{equation}
The mean $\bm \mu_{t, \Delta o}$ and a diagonal covariance matrix $\bm \Sigma_{t, o}$ are predicted by the forward model $f_{\mathrm{FM}}$. 
Equivalently, for the distribution over actions $p_{\bm \theta}(\hat{\bm a}_{t+1}| \bm o_{t+1}, \bm h_{t})$ the inverse model $f_{\mathrm{IM}}$ outputs a mean $\bm \mu_{t, a}$ and diagonal covariance matrix $\bm \Sigma_{t, a}$. 
We model the probability distribution as
\begin{equation}
    p_{\bm \theta}(\hat{\bm a}_{t+1}| \bm o_{t+1}, \bm h_{t}) = \mathcal{N}( \bm{\mu}_{t, a},  \bm{\Sigma}_{t, a}).
\end{equation}

We can train the overall architecture to minimize the negative log likelihood (NLL) loss. 
The overall loss of our forward-inverse model is defined as
\begin{align}
\begin{split}
    \mathcal{L}_{\bm \theta} = \mathbb{E} \Big[ \sum_t & - \log\big(p_{\bm \theta}(\bm o_{t+1}| \bm o_t, \bm a_t, \bm h_t) \big)  \\
    & \underbrace{- \log \big( p_{\bm \theta}(\bm a_{t+1}| \bm o_{t+1}, \bm h_{t}) \big) }_\textbf{prediction loss}+ \underbrace{ \lambda \Theta (\Lambda_t)}_\textbf{regularization} \Big] \label{eq_lossFIM}. 
\end{split}
\end{align}
This corresponds to the original loss from \method (\eqn{eq_gatel0rdloss}) with the task-based loss $\mathcal{L}_{\mathrm{task}}$ being the summed NLL of actions and observations (prediction loss).
In practice, we use the recently proposed $\beta$-NLL loss\footnote{The $\beta$-NLL loss scales the NLL gradients by a $\beta$-exponentiated per-sample variance for a hyperparameter $\beta \in [0, 1]$. $\beta=0$ corresponds to the normal NLL loss. For $\beta=1$ the mean $\bm \mu$ is learned as when using MSE loss.} \cite{betaNLL} ($\beta=0.5$), to avoid instabilities during training and achieve high-quality predictions.
We model all model components apart from \method,  \ie $f_{\mathrm{init}}$, $f_{\mathrm{FM}}$, and $f_{\mathrm{IM}}$, as feed-forward neural networks. 

Can we expect that the resulting latent states $\bm h_t$ will resemble event codes?
EST proposes that event models are only updated upon transient prediction errors, or surprise \cite{EST}. 
\Eqn{eq_lossFIM} essentially phrases this idea in the form of a loss function: the system is trained to minimize prediction errors of future sensorimotor inputs.
Thereby, the regularization loss punishes changes in its latent state.
Thus, the system learns to change the latent state only when the transient prediction errors outweigh the regularization loss.
In accordance to TEC, the latent states $\bm h_t$ will encode both sensory- and motor-predictive aspects, because they are  used to predict both the next sensory states (\eqn{eq:fm}) and the next actions (\eqn{eq:im}).

\subsection{Event boundary predictions}

\begin{figure}
    \centering
    \includegraphics[trim={2.2cm 0 0 0},clip, width=\linewidth]{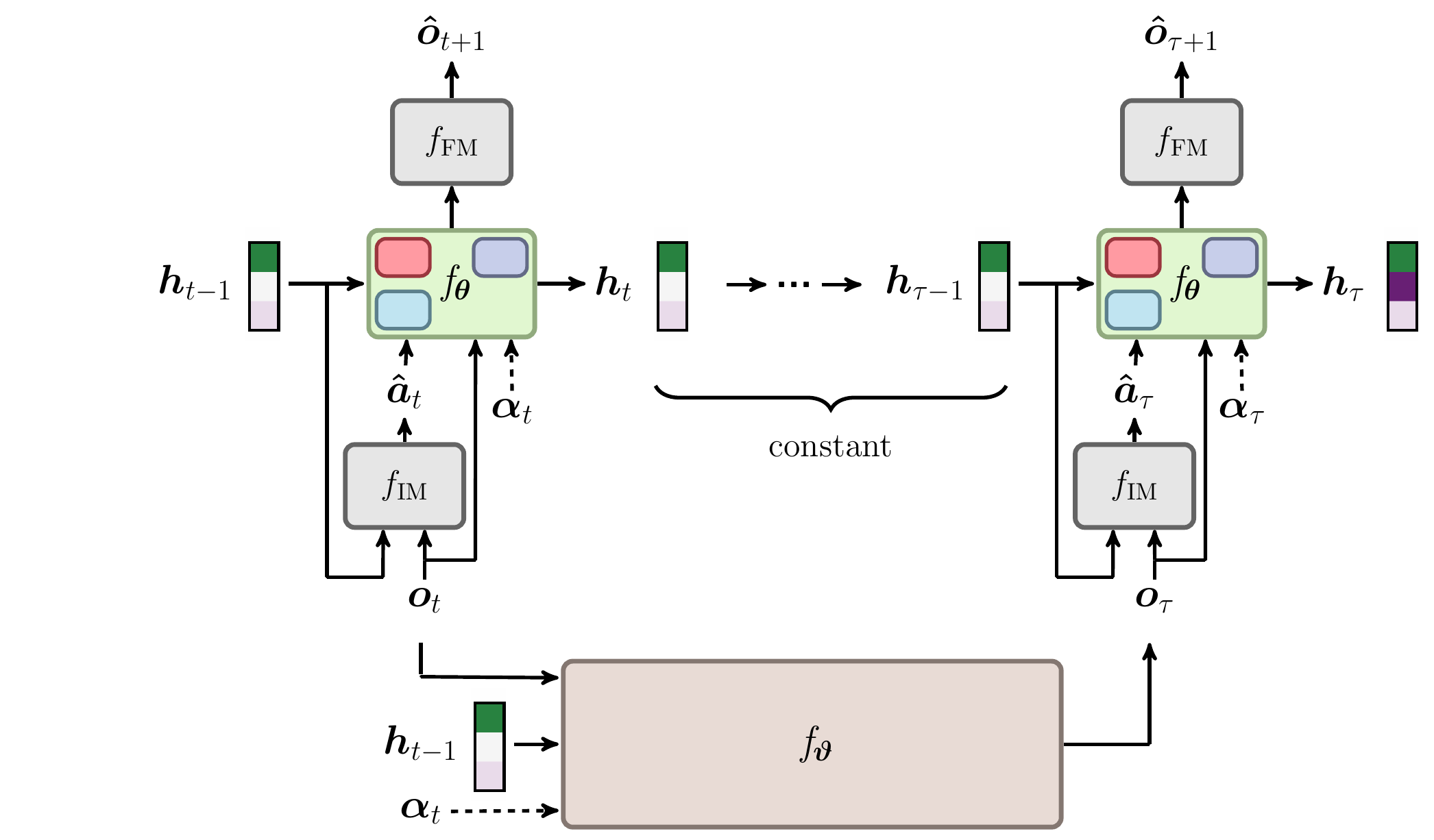}\vspace*{-0.02cm}
    \caption{The overall system: The forward-inverse model $f_{\bm \theta}$ and its latent states $\bm h_t$ are unrolled over time (top). The skip network $f_{\bm \vartheta}$ (bottom) predicts input observation $\bm o_\tau$ for which the latent states change. The attention focus input $\bm \alpha_t$ is only used for modeling infants' goal anticipation (see \sec{sec:modeling_gaze}).}
    \label{fig:skip_net}
\end{figure}

A crucial feature of event cognition is the ability to generate temporally abstract predictions.
We thus add a neural network $f_{\bm \vartheta}$ with learnable parameters $\bm \vartheta$, which we call the \emph{skip network}. 
The skip network is designed to predict the observation that will occur at the next event boundary, that is, at the end of the current event and at start of a next event.
For example the sensation of ones hand touching an object could characterize the ending of a reach-event and the start of a grasp-event.  

Our system should encode an ongoing event with the same latent state $\bm h_t$.
At an event boundary, the latent state changes, which is caused by the opening of an update gate $\Lambda^i_t > 0$ (\eqn{eqn:latent-update}).
We can use this segmentation of the time series to generate the training data for the skip network in a self-supervised fashion. The process is illustrated in \fig{fig:skip_net}.

To generate the training data we feed a sensorimotor sequence $\big((\bm o_1, \bm a_1), (\bm a_2, \bm o_2),..., (\bm o_T, \bm a_T)\big)$ through our forward-inverse model $f_{\bm \theta}$ to receive a sequence of latent states and update gate openings $\big((\bm h_1, \bm \Lambda_1), (\bm h_2, \bm \Lambda_2),..., (\bm h_T, \bm \Lambda_T)\big)$.
We can then determine all event boundaries as
\begin{equation}
    \Tau = \{ t' | \exists_i \Lambda_i^t > 0, 0< t' \leq T\}.
\end{equation}
Thus, $\Tau$ is the set of time steps for which at least one gate opens.
For every time step $t$, we can then determine the next event boundary as
\begin{equation}
    \Tau(t) = \min\big( \{t' \in \Tau | t' > t \} \big).
\end{equation}
We train the skip network to predict $\bm o_{\Tau(t)}$ from the observation $\bm o_t$ and latent state $\bm h_t$.
In other words, the skip network is trained to predict the final observation of an event from an arbitrary starting point within this event.
We treat the last time step T of a sequence also as an event boundary.

To enable the skip network to predict uncertainties we model its output as a distribution of observations $p_{\bm \vartheta}(\hat{\bm o}_{\Tau(t)} | \bm o_t, \bm h_t)$ and use a multivariate Gaussian distribution parametrization:
\begin{equation}
    p_{\bm \vartheta}(\hat{\bm o}_{\Tau(t)} | \bm o_t, \bm h_t) = \mathcal{N}(\hat{\bm o}_{\Tau(t)} | \bm \mu_{t, o_{\Tau(t)}}, \bm \Sigma_{t, o_{\Tau(t)}})
\end{equation}
with mean $\bm \mu_{t, o_{\Tau(t)}}$ and diagonal covariance matrix  $\bm \Sigma_{t, o_{\Tau(t)}}$ predicted by the skip network. The skip network is trained to minimize the loss $\mathcal{L}_{\bm \vartheta}$:
\begin{equation}
    \mathcal{L}_{\bm \vartheta} = \mathbb{E} \Big[ -\log\big( p_{\bm \vartheta}(\bm o_{\Tau(t)} | \bm o_t, \bm h_t) \big) \Big],
\end{equation}
for its learnable parameters $\bm \vartheta$. As before, we use the $\beta$-NLL loss \cite{betaNLL} ($\beta=0.5$).
We model the skip network as an MLP. 

\subsection{Modeling infants' goal anticipation} \label{sec:modeling_gaze}

As previously done in \cite{CAPRI}, we use the structure of events and event boundaries to model experimental findings on infants' goal-predictive gaze.
For modeling infants' gaze (experiments in \sec{sec:gaze_experiments}) we modify the inputs of the two networks.
Here, the forward-inverse model $f_{\bm \theta}$ and the skip network $f_{\bm \vartheta}$ receive an attention focus $\bm \alpha_t$ as an additional input.
We implement the focus as a noise mask on the input observations.
When the system attends to an entity $e$, \eg, the agent's hand, all dimensions of the input observation $\bm o^{e}_t$ concerning this entity receive no sensory noise, e.g. the position of the hand.
All other dimensions of  $\bm{o}_t$, for example the position of an object on a table, are masked by sensory noise.
In this way we model simplified gaze behavior:
The system can focus on entities, akin to looking at them, to gain clear sensory observations, whereas unattended entities receive sensory noise.

Gumbsch et al. \cite{CAPRI} proposed that infants' goal-predictive gaze emerges from selecting gaze behavior that minimizes uncertainty about the future on multiple time scales.
According to their model, infants direct their gaze to minimize uncertainty within the ongoing event to better predict the next sensory observations (intra-event).
Additionally, infants attempt to minimize uncertainty about future event boundaries to better predict when, where, and which future events will follow (inter-event).
In our system we can express this idea as
\begin{align}
\begin{split}
    \bm{\alpha}_t = \argmin_{\bm \alpha} \Big[& \underbrace{U\big( p_{\bm \theta}(\hat{\bm o}_{t+1}| \bm o_t, \bm a_t, \bm h_t, \bm \alpha) \big)}_\textbf{intra-event uncertainty}  \\ &+ \underbrace{U\big( p_{\bm \vartheta}(\hat{\bm o}_{\Tau(t)} | \bm o_t, \bm h_t, \bm \alpha) \big)}_\textbf{inter-event uncertainty}\Big], \label{eq:gaze}
\end{split}
\end{align}
where $U$ is an uncertainty measurement and $p_{\bm \theta}$ and $p_{\bm \vartheta}$ are the Gaussian distributions over observations generated by the forward-inverse-model $f_{\bm \theta}$ and skip network $f_{\bm\vartheta}$, respectively.
We use the following simple uncertainty estimation $U$ 
\begin{equation}
U(p) = \sum_{i \in I} \Sigma_{t,p}^{i,i},  \label{eq:uncertainty}
\end{equation}
which is the sum of variances\footnote{In \cite{CAPRI} entropy was used for the uncertainty $U$. Computing the entropy of Gaussian distributions involves multiplying variances. When some variances are very close to zero, we found the sum of variances to be more stable.}, with $\bm \Sigma_{t,p}$ the predicted covariance matrix of distribution $p$, \ie either $p_{\bm \theta}$ or $p_{\bm \vartheta}$, and $I$ a set of relevant dimensions. $I$ allows us to specify for which parts of the observation the system should aim to minimize uncertainty, \eg, focusing on future hand positions only.

\section{Scenario and data}

We apply our system in the \textbf{Fetch Pick \& Place} environment (OpenAI Gym v1), a benchmark reinforcement learning problem in which a robotic manipulator should move a randomly placed box on a table to a random goal position.
The ``hand'' of the robot is a gripper with two fingers.
The action $\bm a_t \in \mathbb{R}^4$ position-controls the  movement of the hand and the opening of the gripper.
In our experiments the observation $\bm o_t \in \mathbb{R}^{11}$ is composed of the three-dimensional position of three entities, \ie, the hand, the object, and the goal, and the fingers' opening. We modify the simulation such that the height of the table can vary across simulations.

The system is trained and evaluated on two datasets $\mathcal{D}$ of 9.2k sensorimotor sequences  $\big((\bm o_1, \bm a_1), ... (\bm o_T, \bm a_T)\big) \in \mathcal{D}$ of length $T=25$.
The dataset $\mathcal{D}_\mathrm{script}$ contains three types of scripted movements. In  
\textbf{reach-grasp-transport sequences} the hand moves to the object until it is located between the fingers. The hand then closes its fingers and, once the object is fully grasped, lifts the object to the goal position where it is held until the sequence is over.
In \textbf{pointing sequences} the robot ``points'' to the goal by moving its hand to the goal position. 
In \textbf{stretching sequences} the robot stretches its arm by repeatedly performing the same randomly-generated motor command.
In all sequences goal-directed motions follow a fixed direction and decrease their velocity when approaching the target to avoid overshooting.
All actions were perturbed by normal distributed motor noise ($\sigma = 0.05$).

The $\mathcal{D}_\mathrm{APEX}$ dataset contains sequences that were generated by the policy-guided model-predictive control method APEX \cite{APEX} trained to move the object to the goal position.
APEX finds various ways to interact with the object, including lifting, pushing, or flicking it.
In $\mathcal{D}_\mathrm{APEX}$ table height is fixed.

When modeling infants' gaze behavior (experiment \sec{sec:gaze_experiments}), we provide an additional attention focus $\bm \alpha_t$. 
This corresponds to attending to one of the three entities, \ie, hand, object, or goal.
When attending to one entity, sensory information about the other entities' positions are masked by normally distributed noise ($\sigma = 0.05$). 
During training, the attentional focus is randomly shifted 5 times per sequence.

The forward-inverse model and the skip network were trained independently using Adam ($\mathrm{lr}_{\bm{\theta}} = .0005, \mathrm{lr}_{\bm{\vartheta}} = .0001$). Each experiment was run with 10 different random seeds.

\section{Results}
We now first evaluate the semantics of the learned event encodings, then the quality of the skip predictions, the robustness of the learning procedures on more challenging data, and finally the emergence of goal-anticipatory behavior.

\begin{figure}
    \startsubfig{}
  \begin{tabular}{@{}l@{\ \ }l@{\ \ }l@{\ \ }}
    \subfig{fig:scripted_MSE_obs} & \subfig{fig:scripted_MSE_act}&\subfig{fig:scripted_gate_rate}\\[-1em]
   \includegraphics[width=0.325\linewidth]{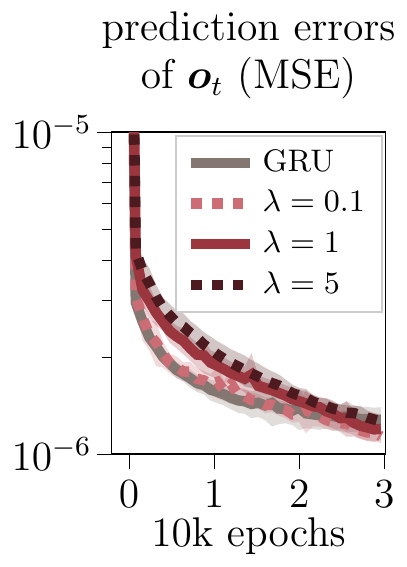}&
    \includegraphics[width=0.325\linewidth]{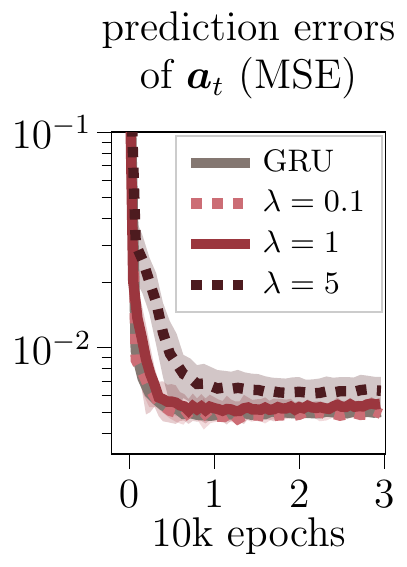}&
    \includegraphics[width=0.325\linewidth]{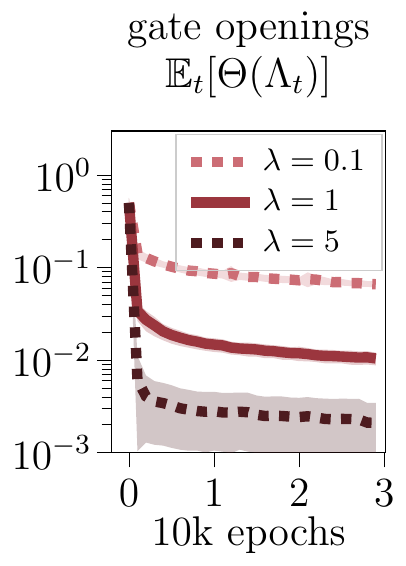}
  \end{tabular}\vspace*{-0.25cm}
  \caption{Results on $\mathcal{D}_\mathrm{script}$: Test prediction error of observations (a) and actions (b) and mean number of gate openings (c), \ie, latent updates. Shaded areas show $\pm$ one standard deviation.}
\end{figure}

\begin{figure*}[h!]
    \startsubfig{}
    
  \begin{tabular}{@{}c@{\ \ }c@{\ \ }c@{\ \ }}
     \subfig{fig:example_reaching} Reach-grasp-transport &  \subfig{fig:example_pointing} Pointing  & \subfig{fig:example_stretching} Stretching\\[-0.5em]
   \includegraphics[width=0.325\linewidth]{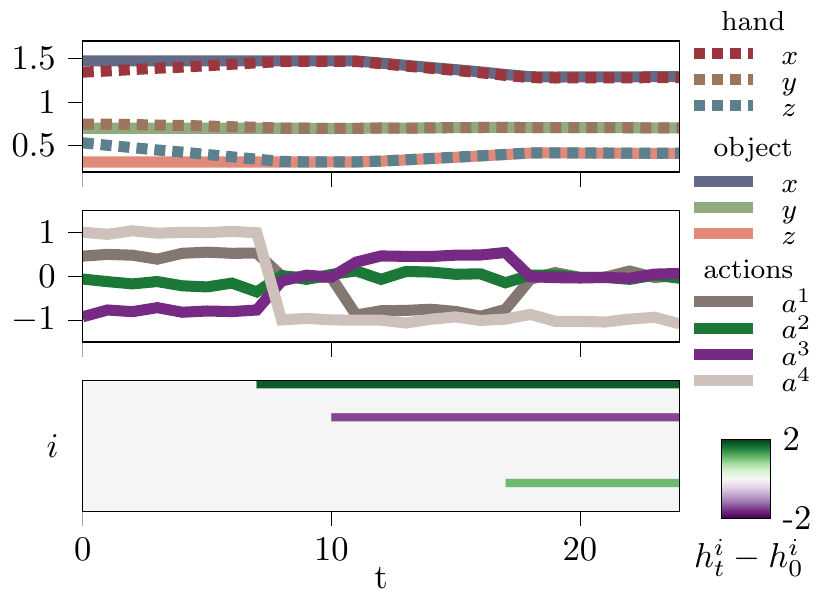}&
    \includegraphics[width=0.325\linewidth]{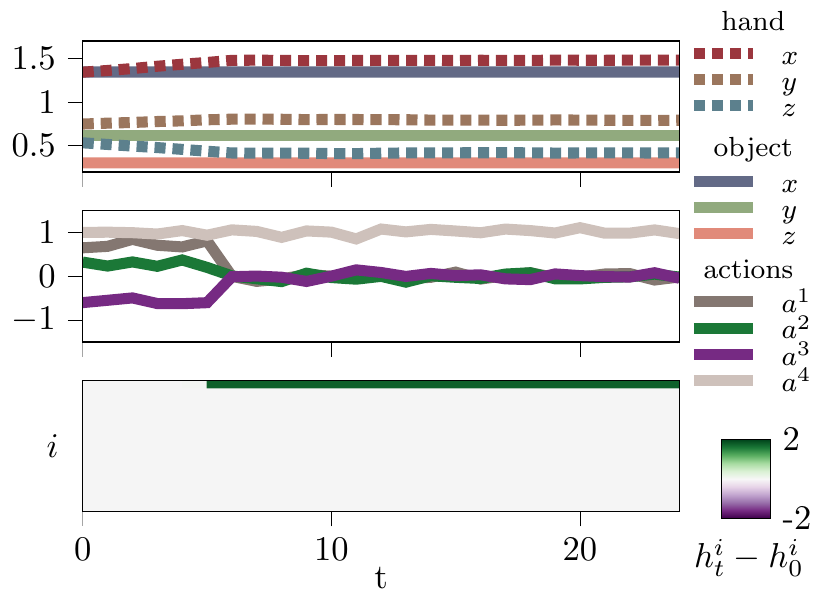}&
    \includegraphics[width=0.325\linewidth]{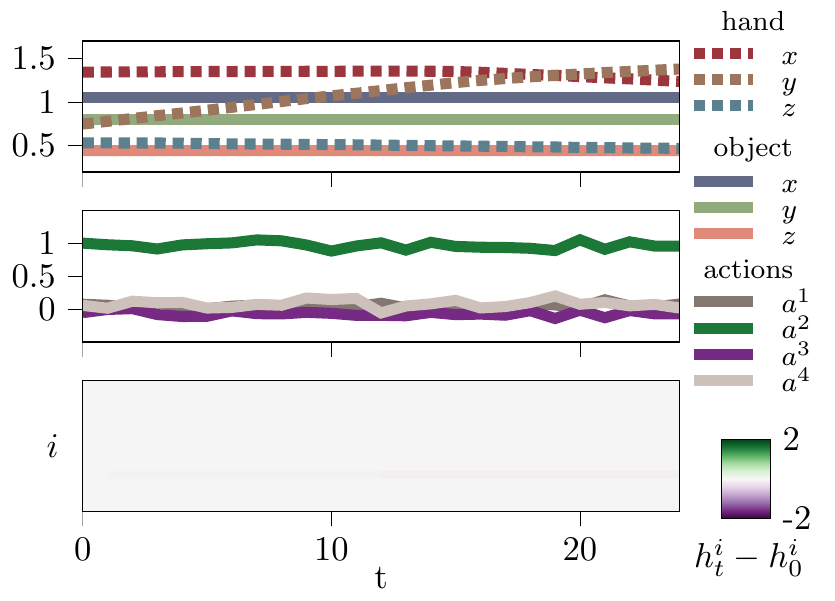}
  \end{tabular}\vspace*{-0.25cm}
  \caption{Exemplary sequences of $\mathcal{D}_\mathrm{script}$ for reach, grasp, and transport of the object (a), pointing to the goal (b), and stretching the arm (c). The topmost row shows hand and object positions over time. The middle row shows actions ($a^{\{1,2,3\}}_t$ control hand movements, $a^4_t$ controls gripper closing). The bottom row shows the latent states relative to initialization, \ie $\bm h_t - \bm h_0$. \label{fig:example_sequences_scripted}}
\end{figure*}

\subsection{Learned event segmentation}

To analyze prediction accuracy and the learned latent state of the forward-inverse model, we first trained our model on the dataset $\mathcal{D}_\mathrm{script}$ of scripted movements.
We compare our system (with \method) with different regularization strengths $\lambda$ to an ablated version without sparsity regularization that uses a GRU \cite{GRU} as its internal RNN cell.

\fig{fig:scripted_MSE_obs} and \fig{fig:scripted_MSE_act} show the prediction errors for the next observations and actions, respectively.
The networks learn to improve their predictions over training.
For a regularization of $\lambda \leq 1$ the regularized versions reach the same prediction accuracy as the GRU ablation.
Thus, the sparsity regularization via \method, does not necessarily degrade performance.
\Fig{fig:scripted_gate_rate} shows the mean number of gate openings, \ie, latent state changes.
With a higher $\lambda$, fewer latent states are updated. 
Our system with $\lambda=1$ learns to only adapt its latent state on average a handful of times during a sequence\footnote{$\mathbb{E}_t\big[\Theta (\bm \Lambda_t)\big] = 10^{-2}$ corresponds to changing one dimension of $\bm h_t$ on average 4 times during a sequence (for $T=25$ and 16-dimensional $\bm h_t$).}, with the same prediction abilities as less regularized variants. 
Thus, for further analysis we focus on our system with $\lambda=1$.

\Fig{fig:example_sequences_scripted} shows exemplary sequences for the three types of scripted movements.
The bottom row illustrates the latent states for one exemplary model.
For the reach-grasp-transport sequence (\fig{fig:example_reaching}), the model changes its latent state  when the hand has reached the object, when the object is fully grasped, and when hand and object arrive at the goal. Thus, the model seems to encode \emph{reaching}, \emph{grasping}, \emph{transporting}, and \emph{holding} the object using four different latent states. 
For the pointing sequence (\fig{fig:example_pointing}), the model changes its latent state once the hand has reached the goal. Interestingly, the same dimension is updated when reaching the object (\fig{fig:example_reaching}) and the goal (\fig{fig:example_pointing}), indicating the similarity between the two events.
For the stretching sequence (\fig{fig:example_stretching}), no clear latent update is visible. This implies that stretching is encoded as one event.

\subsection{Skip predictions} \label{sec_res_skip}
To systematically analyze the skip predictions, we fed the scripted sequences into the forward-inverse model up to a fixed point in time  at $t=2$. 
We then used $(\bm o_2, \bm h_2)$ as inputs to the skip network to produce a skip prediction $\hat{\bm o}_{\Tau(2)}$.
Thus, we essentially queried the skip network at time $t=2$ which observation it predicts at the next event boundary.

\begin{figure}
    \startsubfig{}
    
  \begin{tabular}{c@{\ \ }c@{\ \ }c@{\ \ }}
     \subfig{fig:distance_skip_reaching} Reaching &  \subfig{fig:distance_skip_pointing} Pointing & \subfig{fig:distance_skip_stretching} Stretching\\[0.1em]
   \hspace*{-0.2cm}\includegraphics[width=0.345\linewidth]{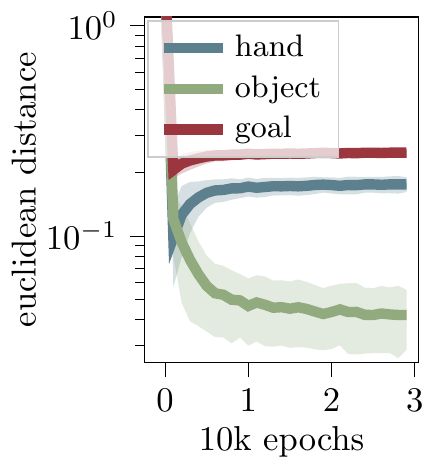}&
    \hspace*{-0.4cm} \includegraphics[width=0.32\linewidth]{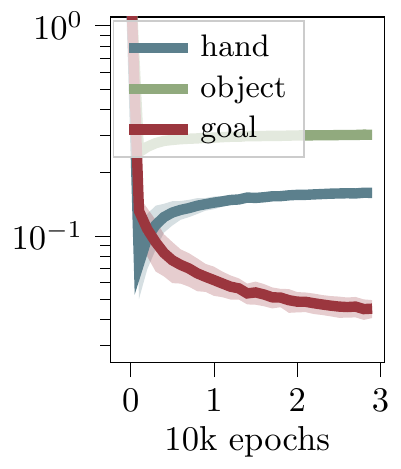}&
    \hspace*{-0.35cm}\includegraphics[width=0.32\linewidth]{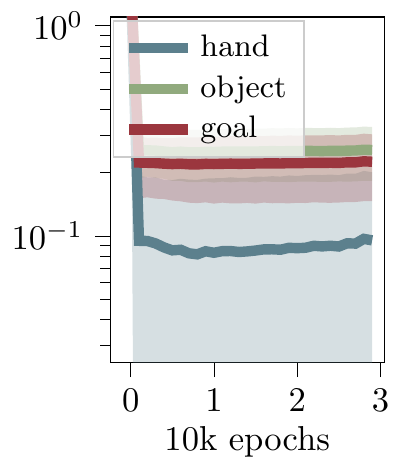}\\
  \end{tabular}\vspace*{-0.25cm}
  \caption{Skip predictions for $\mathcal{D}_\mathrm{script}$. Euclidean distance between skip-predicted hand position in $\hat{\bm o}_{\Tau(2)}$ to the position of all three entities in $\bm o_2$ for a skip at time $t=2$. Shown for reach-grasp-transport (a), pointing (b), and stretching (c) sequences. Shaded areas denote standard deviation. \label{fig:distances_skip}} 
\end{figure}

We compare the predicted position of the hand from $\hat{\bm o}_{\Tau(2)}$ to the current position of the entities in $\bm o_2$.
\Fig{fig:distances_skip} shows the Euclidean distances between the hand position of the skip prediction and the current positions of all entities.
For reaching sequences (\fig{fig:distance_skip_reaching}) the distance between predicted hand position and current object position decreases, while the other distances stay the same or increase.
Thus, the skip network learns that the hand tends to move to the object, where the next gate opening, or event boundary, will occur.
Similarly, for pointing sequences (\fig{fig:distance_skip_pointing}) the system learns that the hand will move to the goal. 
For stretching sequences (\fig{fig:distance_skip_stretching}), on the other hand, the predicted distances between the hand and the other entities show high variances without any regularities.

\begin{figure}
    \startsubfig{}
    
  \begin{tabular}{@{}c@{\ \ }c@{\ \ }c@{\ \ }}
     \subfig{fig:skip_example_reaching} Reaching &  \subfig{fig:skip_example_pointing} Pointing & \subfig{fig:skip_example_stretching} Stretching\\[0.1em]
   \includegraphics[width=0.325\linewidth]{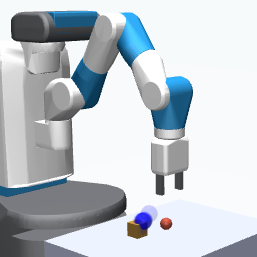}&
    \includegraphics[width=0.325\linewidth]{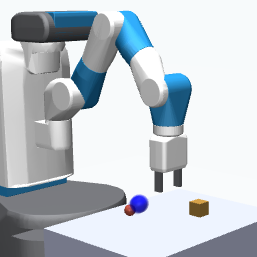}&
    \includegraphics[width=0.325\linewidth]{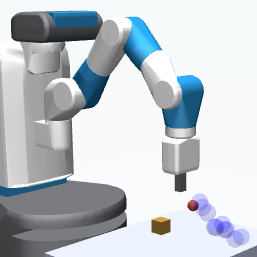}
  \end{tabular}
  \caption{Exemplary skip predictions of the hand. Skips at $t=2$  shown for reaching (a), pointing (b), and stretching (c) sequences of $\mathcal{D}_\mathrm{script}$. One blue circle shows one skip-predicted hand position. Predictions of all 10 random seeds are shown together. The goal position is depicted as a red sphere. \label{fig:skip_examples}}
\end{figure}

\Fig{fig:skip_examples} shows the hand position predictions of all fully-trained skip networks for three exemplary sequences. 
One blue circle marks the predicted position of the hand for one trained skip network.
We show the predictions of all trained networks (10 random seeds) overlayed.
For reach-grasp-transport sequences  (\fig{fig:skip_example_reaching}), the skip network predicts that the hand will be close to the object at the next event boundary.
For pointing sequences (\fig{fig:skip_example_pointing}), the skip network predicts that the hand will be at the goal location (red sphere).
For both sequences the predicted positions strongly overlap, implying a consistent segmentation and hierarchical prediction across different random initializations. 
For stretching sequences (\fig{fig:skip_example_stretching}), the skip network predicts that the hand will be close to positions when the arm is fully stretched-out, or somewhere in-between the current position and the final position.

\begin{figure*}[h]
    \startsubfig{}
    
  \begin{tabular}{@{}l@{\ \ }l@{\ \ }l@{\ \ }l@{\ \ }l@{\ \ }}
     \subfig{fig:apex_MSE_obs} &  \subfig{fig:apex_MSE_act}& \subfig{fig:apex_example} & \subfig{fig:apex_skip_distances}  & \subfig{fig:apex_skip_examples} \\[-1.5em]
   \includegraphics[width=0.157\linewidth]{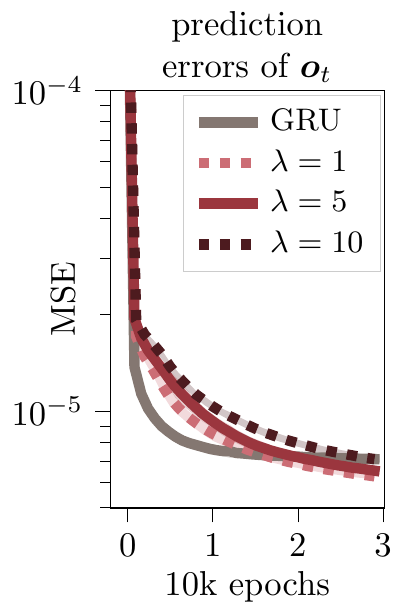}&
    \includegraphics[width=0.157\linewidth]{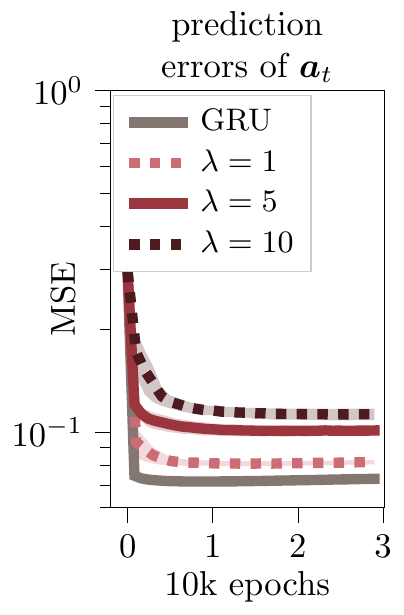}&
    \hspace*{0.1cm}\includegraphics[width=0.32\linewidth]{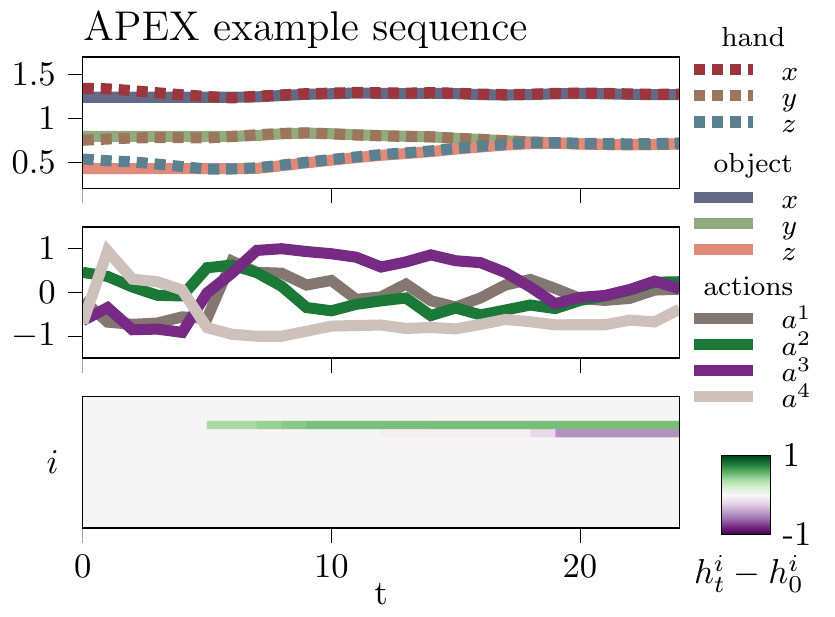} & \includegraphics[width=0.157\linewidth]{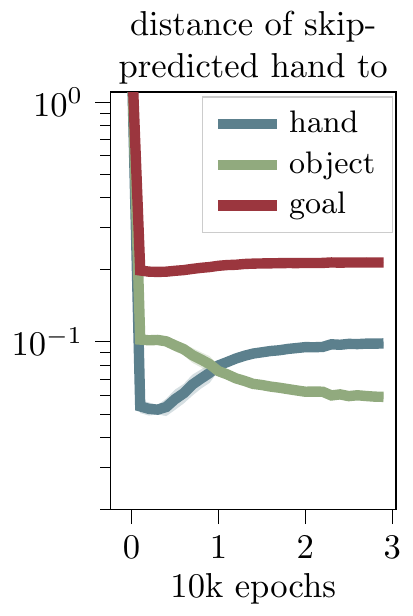} & \hspace*{0.4cm}\raisebox{0.3cm}{\includegraphics[width=0.12\linewidth]{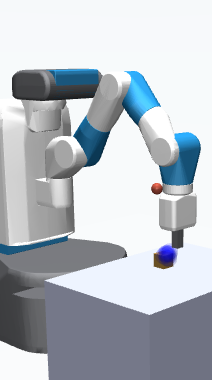}}
  \end{tabular}\vspace*{-0.25cm}
  \caption{Results on $\mathcal{D}_\mathrm{APEX}$: Test prediction error of observations (a) and actions (b). Exemplary reach-grasp-transport sequences (c) showing hand and object position (top), actions (middle) and latent states relative to initialization, \ie $\bm h_t - \bm h_0$ (bottom). Euclidean distance between skip-predicted hand position to the current position of all three entities for a skip at time $t=2$ (d). Exemplary skip predictions (e) at $t=2$ with circles depicting predicted hand positions overlayed for 10 random seeds.}
\end{figure*}

\subsection{Training on more diverse sequences}

To investigate learning robustness, 
we trained our system on sequences of the $\mathcal{D}_\mathrm{APEX}$ dataset, which contains various object interaction sequences. Again, we compare our system (with \method,  $\lambda \in \{1, 5, 10\}$) to a GRU ablation.

\fig{fig:apex_MSE_obs} and \fig{fig:apex_MSE_act} show the prediction errors for the observations and actions, respectively.
While our system outperforms the unregularized GRU ablation for predicting observations, the ablation learns to better predict actions.
For further evaluations we focus on our system with $\lambda=5$.

\Fig{fig:apex_example} shows an exemplary reach-grasp-and-transport sequence from $\mathcal{D}_\mathrm{APEX}$.
The latent states, depicted in the bottom row, changes strongly at two points in the sequence.
First, when the object is grasped, and later, when the object was successfully transported to the goal.
Thus, the latent states seem to encode a similar event structure as for the scripted data (\cf \fig{fig:example_reaching}), even for the much noisier actions of $\mathcal{D}_\mathrm{APEX}$.

To evaluate the system's ability to make temporal abstract predictions, we input  $(\bm o_2, \bm h_2)$ to the skip network and analyze the resulting predictions $\hat{\bm o}_{\Tau(2)}$. 
\Fig{fig:apex_skip_distances} shows the Euclidean distance between the skip-predicted hand position in $\hat{\bm o}_{\Tau(2)}$ and the current positions of all entities at the time $t=2$. 
Over the course of training, the distance between predicted hand position and current object position decreases.
Thus with increasing experience the skip network learns that the hand will move to the object. 
Additionally, the increasing distance between current and predicted hand position suggests that with more training experience the skip predictions reach farther into the future.
\Fig{fig:apex_skip_examples} shows the skip-predicted hand positions for one exemplary sequence. 
Across all random seeds, the system reliably learns to predict that the hand will be close to the object position at the next event boundary.

\subsection{Modeling infants' goal anticipations} \label{sec:gaze_experiments}

\begin{figure*}[b]
    \startsubfig{}
  \begin{tabular}{@{}l@{\ \ }l@{\ \ }l@{\ \ }l@{\ \ }l@{\ \ }}
    \subfig{fig:reach_gaze_gamma1}& \subfig{fig:reach_gaze_gamma0} &\subfig{fig:reach_gaze_gamma05}  & \subfig{fig:point_gaze_gamma05} & \subfig{fig:infant_gaze}\\[-1em]
   \includegraphics[width=0.2\linewidth]{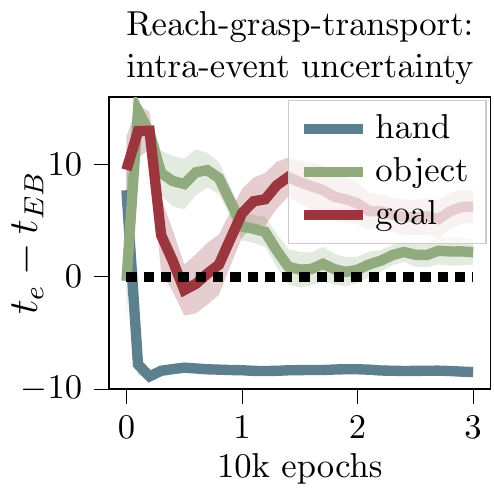}&
   \hspace*{-0.1cm}\includegraphics[width=0.2\linewidth]{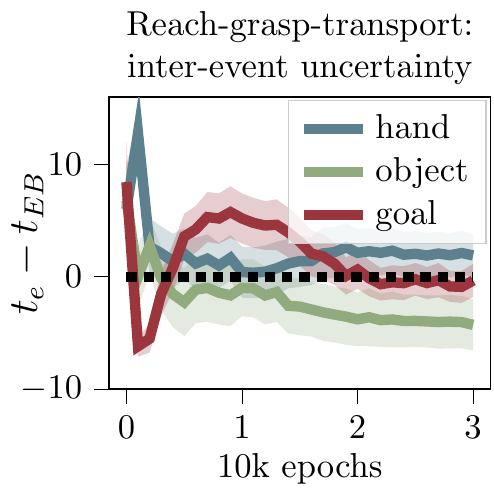}&
   \hspace*{-0.1cm}\includegraphics[width=0.2\linewidth]{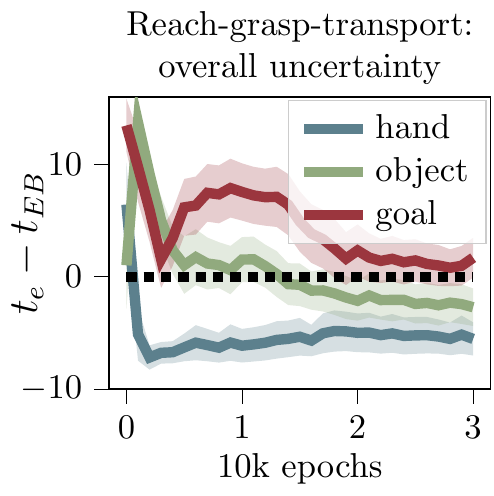}&
   \hspace*{-0.1cm}\includegraphics[width=0.2\linewidth]{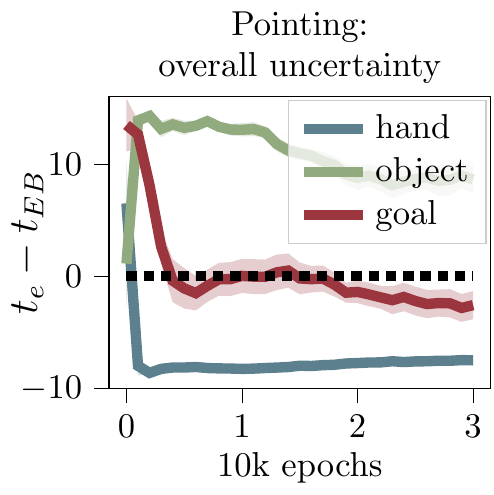}&
   \hspace*{-0.2cm}\includegraphics[width=0.1675\linewidth]{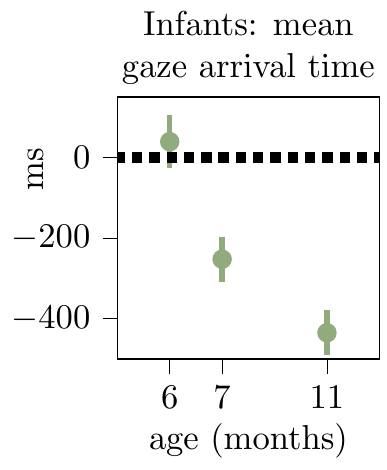}
  \end{tabular} \vspace*{-0.25cm}
  \caption{Goal anticipation of our system and infants. 
  The $y$-axis shows the time $t_e$ of first attending to entity $e$ relative to the time $t_{EB}$ of the first event boundary of a sequence, \ie $t_e - t_{EB}$.
  Thus, negative values for an entity indicate that the system attended to this entity before the first event was over.
  The first event boundaries are marked as dashed black lines (reaching: hand at object; pointing: hand at goal). For reaching events we evaluate attention selection for minimizing uncertainty within the event (a), about the next event boundary (b), or both (c). For pointing sequences we only consider minimizing both uncertainties (d). In (e) the mean gaze arrival time for infants watching grasping sequences are shown (from \cite{Adam2020}). Gaze arrival times below the dashed line (hand object contact) mark anticipatory gaze. Shaded areas and error bars show standard error.}
    \label{fig:gaze_res}
\end{figure*}

Lastly, we model experimental findings of goal anticipations in infants.
We train our system on  $\mathcal{D}_\mathrm{script}$ with an additional attentional focus $\bm \alpha_t$.
We test the system similarly to eye-tracking studies on infant goal anticipation, \eg \cite{Adam2020}.
We show the system observation sequences\footnote{Action inputs are generated by the inverse model $f_\mathrm{IM}$. Only the first action $\bm a_1$ is provided to initialize the latent state $\bm h_0$ and start the process.} and let it choose it's focus $\bm \alpha_t$ to minimize uncertainty about future positions of the hand (specified as indices $I$ in \eqn{eq:uncertainty}).
Similar to how the developmental studies track the first gaze to a goal area, we track the time step $t_e$, when the system first attended to entity $e$ (hand, object, or goal)\footnote{Attention is selected as in \eqn{eq:gaze}. For each entity $e$, we determine the first time step $t_e$ for which $\alpha^e_{t_e} = 1$, with $\bm \alpha_t$ the one-hot encoded attentional focus. No focus on an entity is treated as $t_e = 25$, \ie, maximum length $T$.}.
Akin to infants' goal-predictive gaze shifts, we want to see if our system develops a \emph{goal-predictive attention shift}: Does it attend to the goal of an event, \eg, the to-be-grasped object, before the goal is reached?

We compare three different versions of attention selection for reaching events: Minimizing uncertainty within the current event (intra-event uncertainty in \eqn{eq:gaze}), minimizing uncertainty about the next event boundary (inter-event uncertainty in \eqn{eq:gaze}), or minimizing both uncertainties (full \eqn{eq:gaze}).
When minimizing intra-event uncertainty for reaching (\fig{fig:reach_gaze_gamma1}), the system first attended to the hand (blue) long before hand-object-contact (dashed black line) with little variance across simulations.
Naturally, attending to the hand helps to predict immediate hand movements during reaching. 
The system on average attended to other entities only after hand-object-contact with higher variance across simulations. 
When minimizing inter-event uncertainty (\fig{fig:reach_gaze_gamma0}), the system on average first attended to the object (green) before it was reached by the hand (dashed black line).
Apparently, the system has learned that attending to the object helps to minimize uncertainty about the end of a reaching event.

When putting both uncertainties together, the system exhibits goal-predictive attention shifts: For reach-grasp-transport sequences (\fig{fig:reach_gaze_gamma05}), the system on average first attended to the hand (blue), followed by attending to the object (green). 
The focus on the object happened on average before hand-object-contact  (dashed black line). 
Similarly, for pointing sequences (\fig{fig:point_gaze_gamma05}) the system on average first attended to the hand (blue) and then to the goal (red) before the hand arrived there (dashed line).
Thus, in both cases, the system shifted its attention to the end of the event before the event was over.
Crucially, this behavior developed with training experience.

This behavior mimics gaze-behavior found in infants:
\Fig{fig:infant_gaze} shows the mean gaze arrival time for infants watching movies of a hand grasping and lifting a toy \cite{Adam2020} depending on age.
Mean gaze arrival times were calculated by subtracting the time when the hand entered the target object area of interest (AOI), from the time of the first fixation to that AOI. 
Negative gaze arrival times thus indicate goal-predictive gaze.
Only trials in which the hand was first fixated (for a minimum of 200 ms) were considered.
As shown in \fig{fig:infant_gaze}, 6-month-olds tend to follow the hand with their gaze.
Older infants (7 and 11 months) shift their gaze from the hand to the object before it is reached. 
These results are qualitatively matched by our systems' behavior (\cf \fig{fig:reach_gaze_gamma05}, green line).

\section{Discussion}

We propose a hierarchical deep learning system that learns to encode its sensorimotor experience in event-compressing, latent states, which are used for probabilistic forward- and inverse predictions.
We show that the learned codes can uncover suitable sensorimotor sequence segmentations. 
Additionally, the system can learn meaningful temporal abstract predictions of latent state changes, that is, event boundaries.

Previously, \cite{CAPRI} used an explicit structure of events and their boundaries to model the development of goal-predictive gaze in infants. 
However, in their setup events unfolded linearly and the segmentation was supervised.
Our system learns event segmentation in a completely self-supervised fashion.
This allowed us to apply uncertainty minimizing gaze inference strategies without providing any event information explicitly. 

We tested our model similarly to how goal anticipation was studied in infants \cite{Kanakogi2011, Adam2020}. 
Our system automatically learned to shift its focus from the moving entity (hand) to the next goal as it became more and more mature. In this way it replicates the  development of goal-predictive gaze shifts in infants.
Hypotheses that event familiarity and motor experience enable goal-predictive gaze shifts \cite{Kanakogi2011, elsner2021infants} align with our model.

Besides general neural network related hyperparameters, our model only has one hyperparameter that needs to be set: the regularization strength $\lambda$, controlling the trade-off between prediction accuracy and sparsity in latent state changes. 
The parameter is important for developing robust event segmentation and thus the emergence of meaningful goal-predictive attention shifts.
While  \method works rather robustly under a range of values for $\lambda$, in the future it may be interesting to investigate a battery of gates with their own $\lambda$ values to foster the development of hierarchical event segmentation \cite{Gumbsch:2017}.

Seeing that our model learns event segmentation best when there is clear event-separating sensorimotor information available, our model predicts that salient segmentation cues, \eg, a sound at the event boundary, should ease event segmentation and, thus, boost goal anticipations.
Future infant studies could test whether goal anticipation for newly learned event sequences occurs at an earlier age if during learning of sequences salient segmentation cues are provided.

We believe our system is not limited to modeling goal-predictive gaze.
Event-based anticipations are ubiquitous \cite{Butz2021}.
Moreover, we expect to show that our system is able to plan hierarchically, to exhibit complex one-shot learning, and to solve challenging reasoning tasks \cite{options, Eppe2022}.

\section*{Acknowledgment}

We acknowledge the support from the Machine Learning Cluster of Excellence, EXC number 2064/1--project number 390727645, the Tübingen AI Center (FKZ: 01IS18039B), the Volkswagen Stiftung (No. 98 571), the DFG Priority-Program ``The Active Self'' SPP 2134 (project BU 1335/11-1, EL 253/8-1) and the IMPRS-IS.

\bibliography{mybibfile}
\bibliographystyle{icml2021}

\appendix

\subsection*{A. Implementation details} \label{sec_model_details}
We use \method \cite{GateL0RD} with a 16-dimensional latent state $\bm h_t$. 
The subnetworks $g$ and $r$ are implemented by multilayer perceptrons (MLPs) with three layers ($64 \rightarrow 32 \rightarrow 16$ feature neurons, $\tanh$ hidden activation), whereas $o$ is implemented as two single-layered neural networks ($16$ feature neurons per layer), whose outputs are multiplied element-wise. When using the GRU we found that a 16-dimensional latent state $\bm h_t$ performs much worse on $\mathcal{D}_\mathrm{APEX}$, which is why we used the GRU with 32 latent dimensions.
The other components of the forward-inverse model, \ie $f_{\mathrm{init}}$, $f_{\mathrm{FM}}$, and $f_{\mathrm{IM}}$, are also implemented as feed-forward neural networks.
For $f_{\mathrm{init}}$ and $f_{\mathrm{FM}}$ we use MLPs with three layers (64 $\rightarrow$ 32 $\rightarrow$ 16 feature neurons, $\tanh$ hidden activation).
The inverse model $f_{\mathrm{IM}}$ has an additional mutliplicative layer, akin to the \method output function $o$ (16 features per layer), followed by a three-layered MLP (64 $\rightarrow$ 32 $\rightarrow$ 16 feature neurons , $\tanh$ hidden activation).
We add this extra layer, such that $f_{\mathrm{IM}}$ has the same computational power to compute its outputs based on $\bm o_t$ and $\bm h_t$ as $f_{\mathrm{FM}}$. We model the skip network as a deep MLP with five layers ($512 \rightarrow 256 \rightarrow 128 \rightarrow 64 \rightarrow 32$ feature neurons, $\tanh$ hidden activation). 
Networks predicting Gaussian distributions (\ie $f_{\mathrm{FM}}$, $f_{\mathrm{IM}}$ and $f_{\bm \vartheta}$) use separate read-out layers for predicting the mean and predicting the covariance. The read-out layers for mean predictions have a linear activation function, whereas the read-out layers for the covariance matrix predictions use the ELU activation function shifted by 1 to avoid covariances $\leq 0$.

We trained and tested on datasets of 9.2k sequences using a batch size of 192.
The forward-inverse model and the skip network were trained independently using Adam ($\mathrm{lr}_{\bm{\theta}} = 0.0005, \mathrm{lr}_{\bm{\vartheta}} = .0001, \epsilon=10^{-4}$) and gradient norm clipping (max$=0.1$).
The code to run our experiments can be found at \url{https://github.com/CognitiveModeling/HierarchicalGateL0RD}.

\subsection*{B. Skip predictions of goal and object positions}

In \sec{sec_res_skip} we analyzed the skip predictions and showed that our system learns to make meaningful temporal abstract predictions of future hand positions.
How are the skip predictions for the positions of the other two entities?

\begin{figure}
    \startsubfig{}
    
  \begin{tabular}{@{}c@{\ \ }c@{\ \ }c@{\ \ }c@{\ \ }}
     Reaching &  Pointing &  Stretching & Transport \\[0.1em]
   \includegraphics[width=0.225\linewidth]{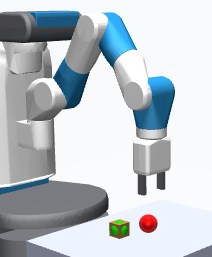}&
    \includegraphics[width=0.225\linewidth]{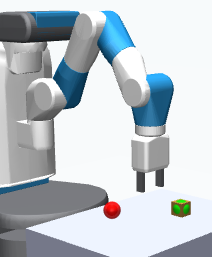}&
    \includegraphics[width=0.225\linewidth]{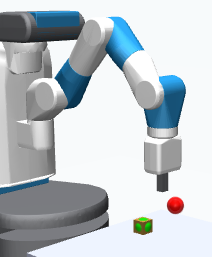}&
    \includegraphics[width=0.23\linewidth]{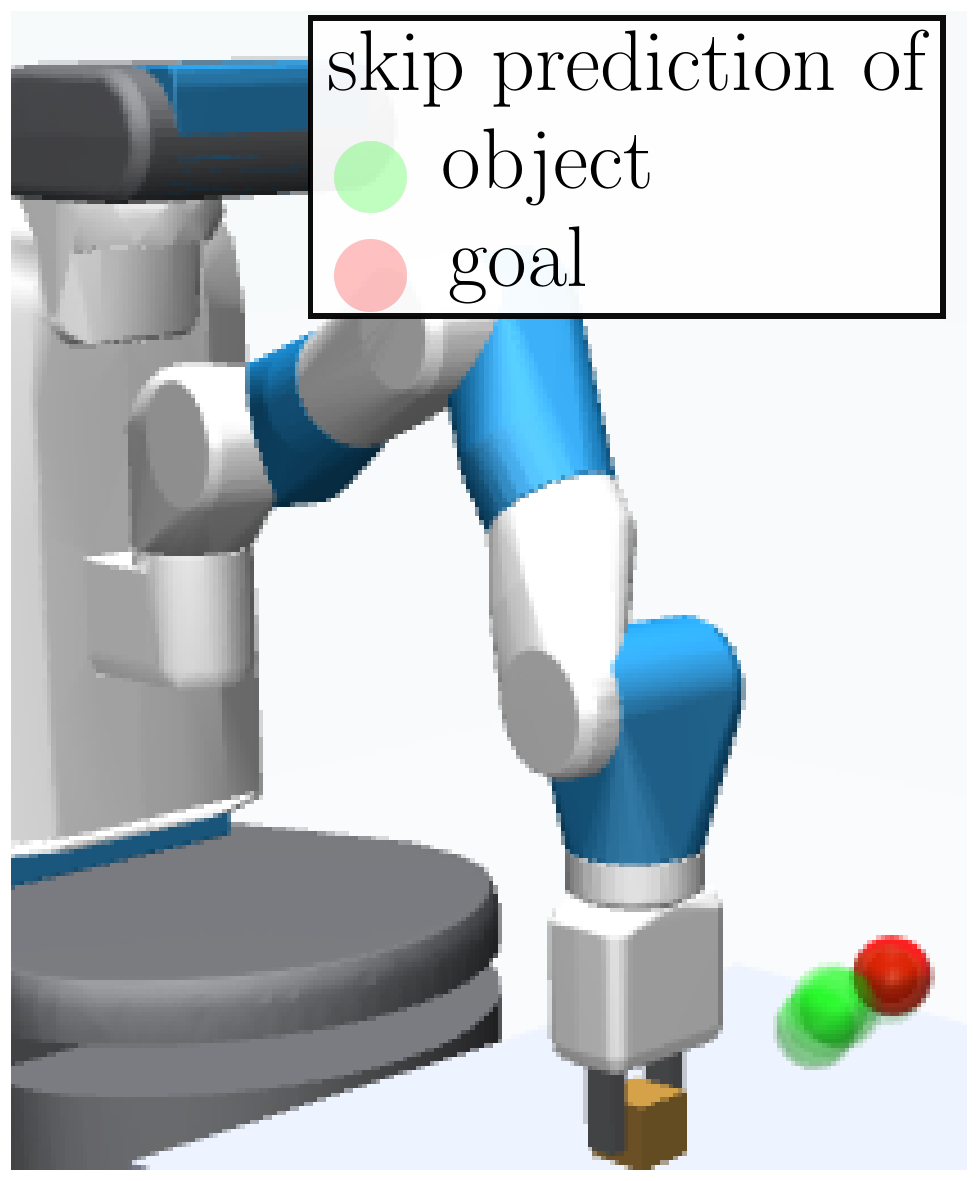}
  \end{tabular}\vspace{0em}
  \caption{Exemplary skip predictions for object and goal positions. Skips at $t=2$  shown for reaching (a), pointing (b), and stretching (c) sequences.  Skips at $t=12$  shown for a  transport event (d). One circle shows one skip-predicted position of the object (green) or the goal (red). Predictions of all 10 random seeds are shown overlayed. \label{fig:skip_examples2}}
  \vspace*{-0.4cm}
\end{figure}

We investigated the skip-predictions for object and goal positions as before by feeding the scripted sequence of $\mathcal{D}_\mathrm{script}$ into the forward-inverse model to generate inputs for the skip network.
\Fig{fig:skip_examples2} (a)-(c) shows the skip predictions for three exemplary sequences based on the inputs $(\bm o_2, \bm h_2)$. 
For all sequences the skip network predicts that the object position (green) and the goal position (red) will not change (\cf \fig{fig:skip_examples} for hand predictions and the ground truth goal positions).
This is reasonable for the given events, since the goal position never changes and for reaching, pointing, and stretching events the positions of the object does not change.
However, for reach-grasp-transport sequences, the object position changes once the object is grasped (typically at $t \leq 12$).
As shown exemplary in \fig{fig:skip_examples2} (d), if we used $(\bm o_{12}, \bm h_{12})$ as inputs to the skip network for a reach-grasp-transport sequence, the skip network indeed predicts that the object will be close to the goal at the next event boundary.
Thus, the skip network can also make temporal abstract predictions of the object and the goal position.

\end{document}

%% file: header.tex
\usepackage{graphics}
\usepackage{mathtools}

\usepackage{algorithm}
\usepackage{algorithmic}
\usepackage{enumitem}

\usepackage{amsfonts}       % blackboard math symbols
\usepackage{amsmath}       % blackboard math symbols
\usepackage{amssymb}
\usepackage{amsthm}
\usepackage{bm}

\newcommand{\Fig}[1]{Figure~\ref{#1}}  % beginning of sentence
\newcommand{\fig}[1]{Fig.~\ref{#1}}    % somewhere

\newcommand{\Eqn}[1]{Equation~\ref{#1}}
\newcommand{\eqn}[1]{Eq.~\ref{#1}} % equation 1.1
\renewcommand{\sec}[1]{Sec.~\ref{#1}} % Section 1

\usepackage{xspace}
\makeatletter
\DeclareRobustCommand\onedot{\futurelet\@let@token\@onedot}
\def\@onedot{\ifx\@let@token.\else.\null\fi\xspace}
\def\eg{e.g\onedot}
\def\ie{i.e\onedot}
\def\cf{cf\onedot}

\makeatother

\DeclareMathOperator*{\argmin}{arg\,min}

\let\originalleft\left
\let\originalright\right
\renewcommand{\left}{\mathopen{}\mathclose\bgroup\originalleft}
\renewcommand{\right}{\aftergroup\egroup\originalright}

\newcommand{\g}[1]{% Growing parenthesis
  \ifthenelse{\equal{#1}{(}}
  {\left( }%
    { \ifthenelse{\equal{#1}{)}}
      { \right)}%
    { \ifthenelse{\equal{#1}{[}}
      {\left[}%
        { \ifthenelse{\equal{#1}{]}}
          { \right]}%
        {#1}}
    }
  }
}

\newcounter{mysubfig}%
\newcounter{myfig}%
\newcommand{\startsubfig}{\setcounter{mysubfig}{0}\setcounter{myfig}{\value{figure}}\addtocounter{myfig}{1}\small }%
\newcommand{\subfig}[1]{\refstepcounter{mysubfig}\label{#1}(\alph{mysubfig})\xspace}%

\usepackage{xcolor}

\definecolor{chrisblue}{RGB}{99,106,134}
\definecolor{chrisgreen}{RGB}{145,170,126}
\definecolor{chrisred}{RGB}{155,54,62}

\definecolor{ourorange}{rgb}{0.881,0.611,0.142}
\definecolor{ourviolet}{rgb}{0.528,0.471,0.701}
\definecolor{ourbrown}{rgb}{0.772,0.432,0.102}
\definecolor{ourlightblue}{rgb}{0.364,0.619,0.782}
\definecolor{ourdarkgreen}{rgb}{0.572,0.586,0.}